\documentclass[sn-mathphys]{sn-jnl}
\usepackage{mathptmx}
\usepackage{graphics}
\usepackage{amssymb}
\usepackage[utf8]{inputenc}
\usepackage{makecell}
\usepackage{multirow}
\usepackage{hyperref}
\usepackage{newtxtext,newtxmath}

\jyear{2021}%

\theoremstyle{thmstyleone}%
\theoremstyle{thmstyletwo}%

\theoremstyle{thmstylethree}%
\newtheorem{definition}{Definition}%

\begin{document}

\title[Federated Few-shot Learning for Cough Classification with Edge Devices]{Federated Few-shot Learning for Cough Classification with Edge Devices}


\author[1]{\fnm{Ngan} \sur{Dao Hoang}}\email{dhoangngann@gmail.com}

\author[2]{\fnm{Dat} \sur{Tran-Anh}}\email{dat.trananh@tlu.edu.vn}

\author[3]{\fnm{Manh} \sur{Luong}}\email{v.manhlt3@vinai.io}

\author[1]{\fnm{Cong} \sur{Tran}}\email{congtt@ptit.edu.vn}

\author*[1]{\fnm{Cuong} \sur{Pham}}\email{cuongpv@ptit.edu.vn}

\affil[1]{\orgdiv{Faculty of Information Technology}, \orgname{Posts and Telecommunications Institute of Technology}, \orgaddress{\street{Nguyen Trai, Ha Dong}, \city{Hanoi}, \postcode{12100}, \country{Vietnam}}}

\affil[2]{\orgdiv{Faculty of Information Technology}, \orgname{Thuyloi University}, \orgaddress{\street{Dong Da}, \city{Hanoi}, \postcode{116830}, \country{Vietnam}}}

\affil[3]{\orgname{VinAI Research}, \orgaddress{\street{Technopark building}, \city{Hanoi}, \postcode{12100}, \country{Vietnam}}}



\abstract{Automatically classifying cough sounds is one of the most critical tasks for the diagnosis and treatment of respiratory diseases. However, collecting a huge amount of labeled cough dataset is challenging mainly due to high laborious expenses, data scarcity, and privacy concerns. In this work, our aim is to develop a framework that can effectively perform cough classification even in situations when enormous cough data is not available, while also addressing privacy concerns. Specifically, we formulate a new problem to tackle these challenges and adopt few-shot learning and federated learning to design a novel framework, termed F2LCough, for solving the newly formulated problem. We illustrate the superiority of our method compared with other approaches on COVID-19 Thermal Face \& Cough dataset, in which F2LCough achieves an average F1-Score of 86\%. Our results show the feasibility of few-shot learning combined with federated learning to build a classification model of cough sounds. This new methodology is able to classify cough sounds in data-scarce situations and maintain privacy properties. The outcomes of this work can be a fundamental framework for building support systems for the detection and diagnosis of cough-related diseases.}





\maketitle

\section{Introduction}\label{sec:introduction}

\subsection{Background and Motivation}
Respiratory illnesses are highly prevalent, wherein cough serves as a typical and initial manifestation of both asthma and chronic obstructive pulmonary disease (COPD) \cite{corrao1979chronic}, \cite{smith2006cough}.  Recently, the world has revealed the outbreak of Coronavirus disease (Covid-19) which is announced as a global pandemic by the World Health Organization (WHO).\footnote{Source: https://www.who.int/health-topics/coronavirus\#tab=tab\_1, accessed: April 29, 2022.} Covid-19 has lead to 509,531,232 confirmed cases and 6,230,357 deaths.\footnote{Source: https://covid19.who.int/, accessed: April 29, 2022.}  A persistent cough is a common symptom among patients diagnosed with COVID-19 as well as those with respiratory illnesses. Prompt and accurate diagnosis of respiratory diseases through coughing symptoms is critical for healthcare professionals in providing appropriate treatment to patients. Consequently, there has been a significant increase in research aimed at utilizing cough sounds to diagnose respiratory diseases, including cough counting, sound classification, and detection. The current study delves into the challenge of classifying cough sounds in the presence of limited labeled cough data, while ensuring the protection of patients' confidential information.

In recent decades, machine learning algorithms have demonstrated their efficacy in supporting healthcare systems as evidenced by various studies \cite{liao2021recognizing}. There has been a proliferation of research that leverages machine learning models for the detection of coughs, with many studies focusing on analyzing cough sounds to develop cough recognition and classification systems. With the advancements in deep learning, methods based on deep neural networks for the detection and diagnosis of respiratory diseases from lung and cough sounds have been proposed and have shown promising results \cite{lella2022automatic, pham2021cnn}.

However, deep learning models require a significant amount of labeled data for training. This presents a challenge in the case of cough classification, as the collection of labeled cough data is both expensive and risks the disclosure of sensitive information. The former challenge is due to the fact that deep learning algorithms typically require a large amount of data to perform optimally, while the latter challenge arises from the scarcity of correctly labeled cough data. Pathological cough data is only available from patients with the disease and requires expert labeling, making the collection of such data a major challenge in the healthcare field. This is attributed to the need for assistance from trained medical professionals, the protection of privacy, and obtaining consent from patients suffering from diseases \cite{ijaz2022towards}.

Furthermore, the collection of medical data presents two practical challenges. The first challenge is the scarcity of data available in the early stages of novel diseases, as evidenced by the lack of medical data on COVID-19 prior to its outbreak. The second challenge pertains to the importance of security and privacy in the collection of medical data. Data protection laws are implemented to prevent the unauthorized collection of private information, and the use of public devices to record cough sounds is discouraged due to the potential for sensitive patient information to be leaked during data collection. As a result, privacy-preserving data collection has become a topic of significant interest among researchers.

\begin{center}
    \begin{figure}[h]
    \begin{center}
    \includegraphics[scale=0.4]{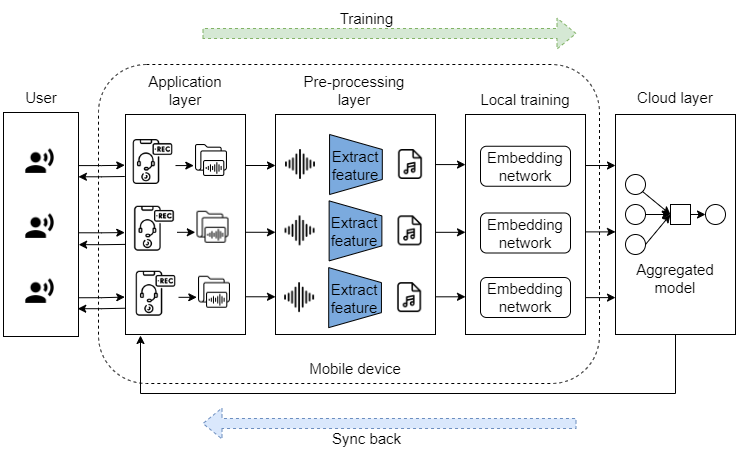}
    \caption{F2LCough system}
    \label{fig:F2LCough system}
    \end{center}
    \end{figure}
\end{center}

The two aforementioned problems inspire us to give birth to an edge-computing framework that supports the classification of novel cough sounds without collecting a large number of data while still ensuring the security of user data. As illustrated in Fig.\ref{fig:F2LCough system}, client data is processed at the periphery of the network, consisting of mobile devices including mobile phones and laptops. To tackle the data scarcity problem, we base on few-shot learning (FSL), which has been receiving a lot of attention \cite{wang2022semantic, bi2022critical} recently. FSL allows us to train a model from an available large dataset, so-called the base set. Then, this model is used to classify novel classes that have not been seen in training while requiring only a few samples of each of novel classes. To address the data security problem, we investigate federated learning, which enables training of models using data located on different local devices owned by local medical centers while effectively protecting medical and health data privacy through parameter exchange. To the best of our knowledge, there has been no prior study of both few-shot and federated learning on cough sounds yet.

\subsection{Main Contributions}
In this study, we aim to to develop a method that effectively learns from limited data while maintaining the privacy of patients who provide audio recordings. To achieve this, we employ a combination of few-shot and federated learning techniques. The study begins by preprocessing the cough audio using Mel Frequency Cepstral Coefficients (MFCCs). The resulting MFCC features are then fed into an embedding network based on ResNet-18 and augmented with an attention mechanism to create embedded vectors, with each class represented by a prototype. The classification task is performed by computing the distances to these prototype representations. Additionally, since the data is stored on multiple devices across various medical centers, the learning of the prototype embedding is conducted in a federated manner, where only the training weights are shared.
Our contributions are summarized as follows:

\begin{itemize}
    \item We formulate a new problem to tackle the cough classification task in the case when data are not only  scarce but also required to be  protected;
    
    \item We propose a novel framework, termed F2LCough, that combines FSL with federated learning techniques to solve the newly formulated problem;\footnote{The source code and data of our proposed framework are available online at  https://github.com/ngandh/F2LCough.} 
    
    \item We empirically evaluate the performance of our proposed framework and competitive baselines using real-world medical data;
        
    \item We evaluate the proposed framework in a real-world scenario, in which the proposed model is deployed on devices with limited computational resources.
\end{itemize}

\subsection{Organization}
The outline of this paper is as follows. In Section \ref{sec:relatedwork}, we review related works for respiratory diseases, coughs, and the ability of FSL combined with federated learning. In Section \ref{sec:methodology}, we describe the definition of the main problem and present our proposed framework. The experimental details and results are presented in Section \ref{sec:experiments}. Finally, the discussion and conclusion are included in Section \ref{sec:conclusion}. 

\section{Related Work}
\label{sec:relatedwork}

\textbf{Cough-related problem.} Detecting cough sounds is an important problem in supporting the diagnosis and treatment of respiratory diseases. A series of research focused on tackling cough classification and cough detection \cite{hoyos2018efficient, Windmon2019TussisWatchAS}. By analyzing unique features of cough sound, the authors in \cite{chatrzarrin2011feature} proposed a method to classify dry and wet cough by utilizing the energy and cough frequency spectrum of cough sounds. However, this approach depends on limited features of sound and thus does not generalize diverse cough sounds in reality. Another method \cite{schroder2016classification} adopted a Gaussian Mixture Model and a Hidden Markov Model (GMM-HMM) to classify dry cough, positive cough, and unknown (arbitrary sounds out of cough sounds). Since GMM faces a distribution problem of data, the proposed method is also not effective in the case of diversified cough sounds. For feature extraction, MFCCs have been used in discriminating between dry coughs and wet coughs \cite{swarnkar2013automatic} as well as classifying tuberculosis coughs. 

Owing to the development of deep learning, several works proved that Convolutional Neural Networks (CNNs) are well-fitted for tasks related to respiratory sounds. Authors in \cite{shuvo2020lightweight} extracted the hybrid scalogram features utilizing the empirical mode decomposition (EMD) and continuous wavelet transform (CWT) of lung sounds, following by a lightweight convolutional neural network to classify respiratory diseases using ICBHI 2017 lung sound dataset \cite{rocha2017alpha}.  For cough identification, the authors in \cite{9720192} proposed a method that used a triplet network architecture that  maps Mel spectrograms of cough sounds to an embedding space via CNN. Another study on Covid-19 cough classification showed that ResNet-50 can discriminate between the COVID-19 positive and healthy coughs with an area under the ROC curve (AUC) of 0.98 \cite{pahar2021covid}. Following up, there is a variety of studies \cite{yang20e_interspeech}, \cite{lella2022automatic} using CNN-based architecture for respiratory tasks effectively. 
Recently,  a series of studies have proven that attention-based modules can enhance the abilities of neural networks on feature recognition \cite{ren2019attention}, \cite{zhao2019exploring}. Zijiang Yang et al. \cite{yang20e_interspeech} proposed a method using ResNet-18 combined with attention blocks to classify adventitious respiratory, which utilized STFT Spectrogram as preprocessed input. 

To be able to deal with deploying models on smart devices, Daniyal Liaqat et al. proposed a CNN-based architecture called CoughWatch which is integrated with smartwatches to detect coughs \cite{liaqat2021coughwatch}. Additionally, a smart-phone based system named TussisWatch \cite{Windmon2019TussisWatchAS} was designed to record and process cough sound for the identification of Chronic Obstructive Pulmonary Disease (COPD) and Congestive Heart Failure (CHF) using Random forests.

\textbf{Few-shot learning.} While deep learning-based methods require a large amount of data to be effective, collecting audio data of people is challenging. Recently, meta-learning and FSL have proven highly effective when labeled data is scarce. In \cite{garcia2018fewshot}, FSL is based on a convolutional network to extract data features and a graph neural network to calculate the classification probability. Based on the theory that FSL is learning to compare samples with each other, a relevant network method was proposed \cite{sung2018learning}. One of the most significant works is the prototypical network \cite{snell2017prototypical}, which is based on the idea that there exists a space embedding in which points are clustered around a unique archetypal representation for each class. 

There have been several studies on FSL for audio classification. In \cite{wang2021few}, the authors proposed a few-shot model by combining ConvNet along with a prototypical network \cite{snell2017prototypical}. In \cite{parnami2022few}, a temporally dilated CNN architecture was designed to combine a novel embedding function with prototypical network \cite{snell2017prototypical} to solve the keyword spotting problem using only limited samples on the speech dataset. Another approach that used a contrastive loss function to learn a latent space from MFCCs to detect COVID-19 cough was proposed in \cite{bhosale2021contrastive}. Recently, few-shot learning has been used in the healthcare field and achieved significant results \cite{wang2022semantic}.

\textbf{Federated learning.} Google AI researchers proposed federated learning for the first time since 2016 to alleviate data privacy issues. Federated learning has been well-allied to many different domains \cite{ek2022evaluation, zeng2023fedprols}. Fed-Affect \cite{shome2021fedaffect}, a few-shot meta-learning method based on the Federated learning framework that can learn from a few labeled images was proposed in recognizing human facial expressions. In \cite{fan2021federated}, authors designed the first federated few-shot learning framework (F2L) by adopting adversarial learning strategy to create a consistent feature space over the clients and optimize the client models to yield a better representation for unseen data samples. Federated learning has been using in the several medical fields to solve privacy and security issues \cite{9774951, 9855868}.

\textbf{Discussion.} Despite the significant results of previous works in cough classification, these methods need a large dataset to achieve good results. Additionally, no prior study in classifying cough has tackled data scarcity and data security simultaneously. 

\section{Methodology}
\label{sec:methodology}
Our paper focuses on the human cough classification task. To tackle data scarcity and information security problems, we proposed a new F2L approach designed specifically for the cough classification problem, where input sound data is classified with corresponding cough labels. In this section, we define the interest problem, present the data processing progress, and describe our proposed framework to address the problem.

\subsection{Problem Definition}
\subsubsection{Notations and Basic Assumptions}

The notations that are used in this paper are summarized in Table \ref{tab:notation}. These notations are formally defined in the following sections that describe our method and technical details. 

\begin{table}[t]
\caption{Summary of notations}
\label{tab:notation}
\centering
\begin{tabular}{ | c | c| } 
  \hline
  \textbf{Notation}& \textbf{Description}\\ 
  \hline
  $U$ & The number of the local devices \\ 
  \hline
  $u$ & The numerical order of the local device \\ 
  \hline
  $e$ & \makecell{An episode in which a few-shot task is performed \\ by utilizing a sampled mini-batch}\\
  \hline
  $S^u$ & The support set of the $u$-th local device \\
  \hline
  $Q^u$ & The query set of the $u$-th local device\\
  \hline
  $p_c^u$ & representation vector (Prototype) for class $c$ in $S^u$\\  
  \hline
  $\mathcal{A}^u$ & The base set of the $u$-th local device\\ 
  \hline
  $\mathcal{E}^u$ & The novel set of the $u$-th local device\\ 
  \hline
  $\omega$ & Weights of embedding network \\ 
  \hline
  $\omega^*$ & Weights of embedding network after aggregating weights \\ 
  \hline
  $f_\omega^u$ & Embedding model with weights $\omega$ of the $u$-th local device \\ 
  \hline
  $f_{\omega^*}^u$ & Global embedding model with weights $\omega^*$ \\ 
  \hline
  $d(a, b)$ & Euclidean distance of vector $a$ and vector $b$. \\ 
  \hline
\end{tabular}
\end{table}

In this study, we tackle the case in which a new type of disease appears suddenly and medical systems have no records related to the disease data before.  We can take COVID-19 as a particular example.  The data can only be collected from patients suffering from these diseases. Moreover, we deal with the situation where we have multiple local medical centers, each can collect a few samples of a patient's cough data. However, these centers do not want to leak their data to protect patient information. Our goal is to classify and detect a novel type of cough without the need for a huge unified dataset.

To this end, we first assume that each local medical center has a sufficient amount of data on several types of traditional cough related to diseases that have been discovered and treated in the past. The available data, the so-called base set, have been labeled and are denoted as {\color{blue}$\mathcal{A}^u$ = $\{(x_i^u, y_i^u) \vert y_i^u \in Y_{train}\}$,} where $x_i^u$ and $y_i^u$ denote the $i$-th cough audio representation and the corresponding cough type on the $u$-th local device, respectively, $Y_{train}$ includes types of cough that have been gathered during the past years. The base set $\mathcal{A}^u$ acts as the prior knowledge, which can be exploited to learn the novel classes of cough through only a few samples. We also assume that there are $U \in \mathbb{N}$ local devices holding cough data belonging to new types of cough that have not been found before and are not included in $\mathcal{A}^u$. The new dataset including only a few labeled data on the $u$-th device is denoted as {\color{blue}$\mathcal{E}^u = \{(x_j^u, y_j^u) \vert y_j^u \in Y_{test}, y_j^u \notin Y_{train}\}$} where $(x_j^u, y_j^u)$ denotes the data point belonging to the novel type of cough on the $u$-th local device, $x_j^u$ is $j$-th audio records of cough, $y_j^u$ is the corresponding label of $x_j^u$, and $Y_{test}$ includes types of novel cough sounds.

We follow the idea of FSL that utilizes the base set $\mathcal{A}^u$ to train a classifier to classify a novel set $\mathcal{E}^u$. In our work, the FSL procedure follows the episodic learning mechanism \cite{snell2017prototypical}, where an episode $e$, including of a support set and a query set, is formed to perform a few-shot task. Intuitively, the two support and query sets are created to mimic the circumstances encountered during real evaluation. In our cough classification context, the support set is the set that has information about the long-established cough sounds and the corresponding labels, which is the prior knowledge that helps to train the model to predict novel labels of data samples in the query set. This idea is based on human learning, for example, a child can utilize knowledge learned from a few sample images of a drawing animal (the support set) to recognize real-world animals (query set).

In the training procedure, we undertake a specific approach wherein we randomly select $N$ classes from the base set $\mathcal{A}^u$ on each device $u$. Within this selection, we further employ random sampling to draw $K$ support data samples and $V$ query data samples per class, thereby composing an episode denoted as $e$. Consequently, within each episode $e$, every class in the selected $N$ classes is represented by two distinct sets: the support set and the query set, both of which consist of sampled images. The support set is {\color{blue}denoted as \begin{math}{S^u = {\{(s_i^u, y_i^u)\}}^{N \times{K}}_{i=1}} \end{math},} including $K$ different audio data samples $\{s_1^u, s_2^u, \cdots, s_K^u\}$ from $N$ classes $\{y_1^u, y_2^u, \cdots, y_N^u\}$, and $(s_i^u, y_i^u) \subset A^u$.  Based on the support set, the classifier predicts a query set. The query set consists of data samples belonging to labels in the support set and is {\color{blue}denoted as \begin{math}Q^u = {\{(q_i^u, y_i^u)\}}^{M \times {V}}_{i=1} \end{math}.} $Q^u$ consists of $V$ different data points \begin{math}{\{q_1^u, q_2^u, \cdots, q_V^u\}} \end{math} from $M$ classes \begin{math}{\{y_1^u, y_2^u, \cdots, y_M^u\} }\end{math}, and $(q_i^u, y_i^u)\subset \mathcal{A}^u$. As the result, the episodes can be characterized by three key variables: the quantity of classes denoted by $N$ (also referred to as the "ways"), the number of samples per class in the support set denoted by $K$ (termed as the "shots"), and the number of samples per class in the query set denoted by $V$.

After the training process is completed, we evaluate the classifier by using the test set which is the novel set $\mathcal{E}^u$. The support set and the query set are taken from the test set $\mathcal{E}^u$ as the way performed in the training process. In the testing process, the support set includes $(s_j^u, y_j^u)\subset \mathcal{E}^u$. And the query set consists of $(q_j^u, y_j^u)\subset \mathcal{E}^u$. In fact, the query set consists of unlabeled data. We only know that the labels of the data in the query set exist in the support set. We have to use an FSL model to predict the labels of data samples in the query set. In our evaluation process, the support set is provided by hospitals, and the query set can be obtained from the patients.

As medical centers have records of cough-related diseases but do not want to leak them out to protect patient information, we adopt federated learning to ensure the privacy of data as well as the satisfaction of patients. To this end, all $U$ local devices are supplied with models that have the same architecture. Each device trains the model with its own data. After that, local devices exchange the weight of models together to produce a global model without sharing data. Specifically, each device learns a local model $f_\omega^u$ following the prototypical network method. After a fixed period, we aggregate local model $f_\omega^u$ from $U$ local devices to generate global model $f_{\omega^*}$, which is called a communication round. After that, the global model is synced back to local devices to continue learning. Finally, we stop the training when we get a good global model $f_{\omega^*}$ that meets our expectations. In this work, we assume a communication round which is performed after all the local devices complete one training round.

\subsubsection{Problem formulation} 
\begin{definition} [The few-shot task of cough classification on the $u$-th local device]\label{def1}
Given labeled data set called a support set $S^u$, few-shot task $\tau^u$ predicts the labels of data points on a query set $Q^u$.
\begin{equation} \label{eq:few-shot task}
{\tau^u(S^u, Q^u) = Y_Q^u}
\end{equation}
where $\tau^u$ is a few-shot task of the support set $S^u$ and the query set $Q^u$. The output of the few-shot task $\tau^u$ is $Y_Q^u$ which is the predicted label set of $Q^u$.
\end{definition}
Definition {\upshape\ref{def1}} formulates the problem that each local medical center can perform few-shot tasks, in which a different embedding space is learned in each local device. 
Specifically, on each device, the objective is to minimize the distance $d(f_\omega^u(q_i^u)$, $p_{c}^u$) at each training step, where $d(\cdot)$ is the Euclidean distance function, $c$ is the label of $q_i^u$, and $p_{c}^u$ is a representation vector called prototype. Each device uses $S^u$ and $Q^u$ to train $f_{\omega}^u$. First, from $f_\omega(s_i)$ of classes, a few-shot model calculates $p_{c}^u$ for each class. Second, the model chooses $c$ that has the smallest distance $d(f_\omega^u(q_i^u)$, $p_{c}^u$)  as the label of query $q_i^u$.

To train a single few-shot task, the objective of each local device  is expressed as:

{\color{blue}
\begin{equation} \label{eq:user loss}
l^u_{\tau^u}(\omega) = -\sum_{i=1}^{Q^u}logP_\omega(y_i^u \vert q_i^u, S^u),
\end{equation}}
where $l^u_{\tau^u}(\omega)$ is the negative log-likelihood of the true label of each query sample ($q_j^u$, $y_j^u$) that the model of the $u$-th local device needs to minimize in a few-shot task ${\tau^u}$.

The average of objectives of few-shot tasks on a device $u$ is formulated as:
{\color{blue}
\begin{equation} \label{eq:Loss}
L^u(\omega) = \frac{1}{\beta^u}\sum_{\tau^u \in \beta^u}^{} l^u_{\tau ^u}(\omega)
\end{equation}}
where $L^u(\omega)$ is the average of the objective functions of few-shot tasks per local device. $\beta^u$ is the total number of few-shot tasks per the $u$-th local device.  

To allow information to be exchanged between local devices, we assume that all $U$ devices can join {\em communication rounds} periodically to generate $f_{\omega^*}$, aggregated from learned models $f_{\omega}^u$ of local devices. A communication round starts when each local device uploads its embedding network $f_{\omega}^u$ to the server and ends when all local devices download the global model.
  To be able to get $\omega^*$, we minimize the average of the objectives of all devices
  {\color{blue}
\begin{equation} \label{eq:obj}
\omega^* = \min _\omega L(\omega) = \sum_{u=1}^{U}\frac{\beta ^u}{\beta}L^u(\omega),
\end{equation}
where $\beta$ is the total number of few-shot tasks on all devices and $U$ is the number of local devices.
}

\begin{definition} [Cough classification]\label{def2}
Given the cough data $x_j^u$ of patient $u$, we aim to predict the cough type of $x_j^u$, which is a member of $Y_{test}$.
\end{definition}
To complete the aim, from (\ref{eq:obj}), our method has learned a transformation function to create an embedding space where cough data points are clustered closely around the class centroid, which is computed as the center point of all $x_j^u$ having the same value of $y_j^u$, denoted as $p^u_{novel}$. Therefore, we train a classifier to choose label $y_j^u$ as the cough type of $x_j^u$ by comparing the smallest distance from the embedded vector of  $x_j^u$ to $p^u_{novel}$, i.e., we find $y_j^u$ satisfying the following condition:
\begin{equation} \label{eq:classify}
y_j^u =\mathrm{argmin}_{y_j^u \in Y_{test}}(d(f_{\omega^*}(x_j^u),p^u_{novel}))
\end{equation}

\subsection{F2LCough Framework}
\subsubsection{Overall Procedure}

\begin{figure*}[h!]
    \begin{center}
    \includegraphics[width=\textwidth]{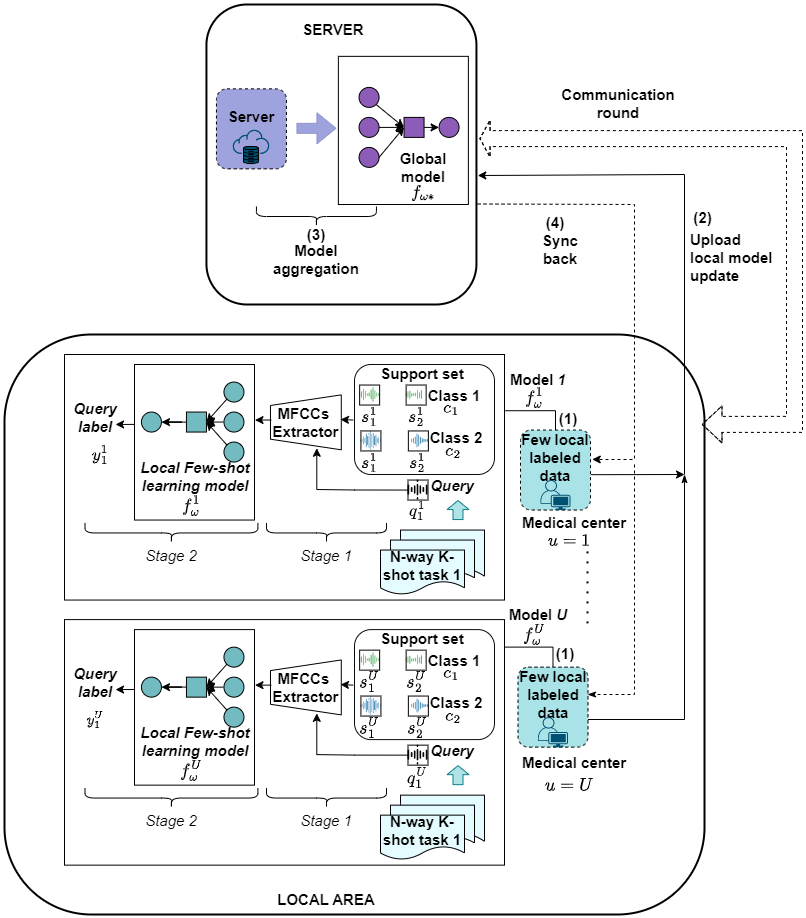}
    \caption{The F2LCough workflow}
    \label{fig:system}
    \end{center}
\end{figure*}

To solve the formulated problem in (\ref{eq:classify}), we propose a framework that consists of two main components: FSL to classify cough sounds and a federated learning paradigm, illustrated in Fig. \ref{fig:system}.

First, we describe the architecture of FSL, where each local device exploits its data to train the embedding model by performing few-shot tasks. To achieve this, we design a training procedure consisting of two stages, as shown in Step (1) in Fig. \ref{fig:system}. The first stage involves extracting Mel Frequency Cepstral Coefficients (MFCCs) from cough sounds in both $S^u$ and $Q^u$. The second stage is dedicated to classifying the query labels, wherein the MFCCs are fed into the embedding network to map the data into the embedding space. This stage utilizes prototypical networks \cite{snell2017prototypical} to create an embedding space for audio records. In this space, classification is performed by calculating the distance to prototype representations of classes.

 In Step (2), all local devices upload their weights ${\omega}^u$ to a server. In Step (3), the global model $f_{\omega^*}$ is aggregated from the uploaded weights. Finally, in Step (4), local devices download and sync the global model to complete a communication round. All four steps are iteratively performed until ${\omega^*}$ is converged.

\subsubsection{Implementation Details}

\textit{Feature Extraction}

Mel Frequency Cepstral Coefficients (MFCCs) are extracted from both support and query data. Empirically, we extract 40-dimensional features of MFCCs and design sliding windows to extract MFCCs in which the length of each window is 128ms with a 64ms hop length. After that, these features are inputted into an embedding network to map data into the embedding space.

\textit{FSL}

\begin{figure*}[h!]
    \begin{center}
    \includegraphics[width=0.6\textwidth]{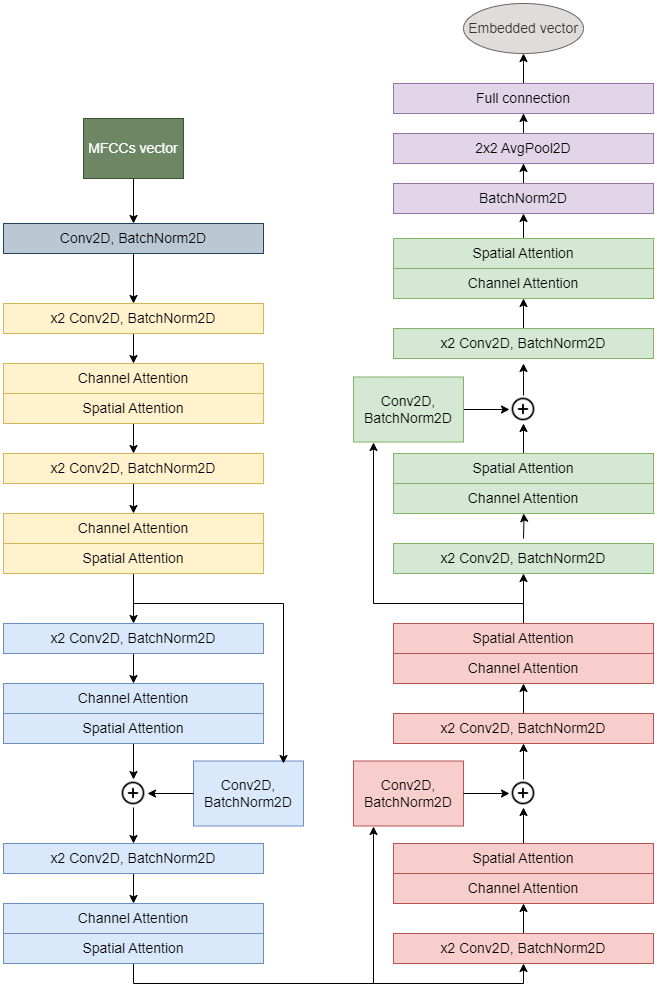}
    \caption{The embedding network}
    \label{fig:embedding network}
    \end{center}
\end{figure*}

The embedding network employed in our study is based on ResNet-18 \cite{he2016deep} architecture, augmented with both channel attention block and spatial attention block. To integrate these components into our embedding network, we follow the approach outlined in \cite{woo2018cbam}. A visual representation of our embedding network is presented in Fig. \ref{fig:embedding network}.

In the embedding space, we follow the implementation of prototypical networks \cite{snell2017prototypical}. For every episode, a prototype is computed by averaging the embedded support examples per class:
{\color{blue}
\begin{equation}
    p_{c}^u = \frac{1}{\vert S_{c}^u\vert} \sum_{(s_{i}^u,y_{i}^u)\in S_{c}^u} f(s_{i}^u)
\end{equation}}
where \textit{f} is embedding function, {\color{blue} $\vert S_{c}^u\vert$  is the number of support examples belonging to class \textit{c} in episode \textit{e}.} 
We apply a softmax function to the negative distance to choose the label of the query sample.


\textit{Federated learning}

We adopt the Federated averaging (FedAvg) algorithm \cite{mcmahan2017communication} to complete the aggregation step, whose objective function is given in (\ref{eq:obj}).



\textit{System design}

To evaluate the proposed framework in a real-world application, edge devices are utilized as in Fig. \ref{fig:F2LCough system}. Specifically, a user has a mobile phone which helps the user to record audio to predict. When training, mobile phones save their data and send it to laptops for training. In the Pre-processing layer, extracting MFCCs of audio data is performed on the laptop in the local training layer. Then, laptops use these MFCCs data to train an embedding network. Each embedding network trained completely in one communication round will be pushed to the cloud layer treated as a server to aggregated weights. Finally, a global model is sent back to the mobile phone to serve users.

\section{Experimental Results}
\label{sec:experiments}

\subsection{Dataset}
\textbf{COVID-19 Thermal Face \& Cough Dataset}: COVID-19 Thermal Face \& Cough Dataset \cite{ward2021flunet} includes the thermal face dataset and the cough dataset. We utilize the cough dataset for our experiments. Each audio file lasts 1 second at the sample rate of 44,100 Hz. The cough dataset consists of  53,471 seconds of “not cough” samples consisting of background noise, office, music, airport, coffee shop, harbor, nightclub, simulated turbulence sounds, and 1,557 seconds of cough sounds. Additionally, 40,856 seconds of cough sounds are augmented with random background noise at a random volume ratio. In our experiments, we only use eight types of cough: \textit{Barking cough, Chesty and wet cough, Coughing up crap again, Dry afternoon cough, Gaggy wet cough, Spring allergy coughing, Heavy cold, and sore throat coughing, Night wet cough}.

\subsection{Baselines}
\label{subsec: baselines}
We compare our proposed method with three state-of-the-art approaches:
\begin{itemize}
    
    \item \textbf{Momentum Contrast (MoCo) \cite{he2020momentum}} is known as a state-of-the-art method of unsupervised learning, it is able to learn a meaningful representation of data for downstream tasks.
    
    \item \textbf{Relational network \cite{sung2018learning}} is a framework for FSL with an end-to-end approach from scratch.
    
    \item \textbf{TC-ResNet}: This model has great results in Keyword Spotting problems \cite{choi19_interspeech}. We utilize a modified version called TD-ResNet7 \cite{parnami2022few}  as an embedding network to compare with our designed embedding network. 
    
    
    
\end{itemize}

\subsection{Experimental Setup}
\label{subsec: setup}


To create a simulation for federated learning, we separate the audio dataset into $U$ = 5 portions, which correspond to 5 local datasets stored in 5 local devices. In each device, we divide the local dataset into two sets: the training set $\mathcal{A}^u$ and the test set $\mathcal{E}^u$. In our experiment, we adopt a cross-validation approach wherein we construct novel sets by selecting two out of the eight types of cough, while the remaining six types are utilized as the base set. Specifically, the novel sets consist of the following pairs: \textit{(Heavy cold and sore throat coughing, Night wet cough)}, \textit{(Dry afternoon cough, Gaggy wet cough)}, \textit{(Spring allergy coughing, Coughing up crap again)}, and \textit{(Chesty and wet cough, Barking cough)}. Additionally, in-depth assessments in this study focus on the novel set containing \textit{(Heavy cold and sore throat coughing, Night wet cough)} due to their closeness to our domain knowledge.

We note that for experiments without federated learning, we solely utilize the data from one user. This approach enables us to evaluate the outcomes when a reduced amount of data is employed.

In the training process, we train our F2LCough with the learning rate set to 0.001 and the number of epochs is 200. 

We utilize the F1-Score as the evaluation metric, which is a widely adopted measure in classification tasks. The results are presented in Mean $\pm$ Standard deviation format and are expressed as percentages.

\subsection{Experimental Results}
Our empirical study in this subsection is designed to answer the following three key research questions.

\begin{itemize}
    \item \textit{Q1.} How effective is our proposed embedding architecture?   
    \item \textit{Q2.} How much can F2LCough boost the security and accuracy of the classifier?
    \item \textit{Q3.} How does F2L Cough operate in practical applications using low-capacity devices?
\end{itemize}

\subsubsection{Comparison between ResNet-18 - Attention and TC-ResNet (Q1)}

Since our proposed embedding network architecture is similar to TC-ResNet, we perform an ablation study that replaces our embedding network, in which ResNet-18 combined with attention mechanisms is employed, with an implementation of TC-ResNet (termed TD-ResNet7) in \cite{parnami2022few}. In this experiment, we only consider the performance comparison in a few-shot setting without federated learning. More specifically, all the few-shot settings are listed as follows: 2-way 2-shot, 2-way 8-shot, and 5-way 8-shot. Table \ref{tab:few-shot} illustrates the performance comparison between TD-ResNet7 and our proposed ResNet-Attention architectures. From the table, we discuss the following interesting observations:
\begin{itemize}
\item We observe that the performance of TD-ResNet7 is slightly superior to our ResNet-Attention in the case of the 2-way 2-shot task. 
\item However, except for the 2-way 2-shot task, the F1-Scores obtained by using TD-ResNet7 are significantly lower than those from ResNet-Attention in the remaining tasks. 
\end{itemize}
Intuitively, although both TD-ResNet7 and our proposed architecture are built upon ResNet-18, the difference in implementation details causes the performance gap between the two architectures. Specifically, TD-ResNet7 sets the kernel sizes to $7\times1$ and $3\times1$ instead of the square kernels in ResNet-18. Besides, TD-ResNet7 applies dilated convolutional layers instead of normal convolutional layers. This architecture of TD-ResNet7 performs effectively on normal sequential data such as speech. However, the cough has a cycle that is not as same as other sequential data; therefore, TD-ResNet7 is unable to capture meaningful features of cough sound while our proposed architecture can do better.

\begin{table*}[t]
\caption{F1-Scores comparison in varied few-shot settings}
\label{tab:few-shot}
\begin{center}
\resizebox{.9\textwidth}{!}{%
\begin{tabular}{ |c|c|c|c|c| }
\hline
\textbf{Label} & \textbf{Model} & \textbf{2-way 2-shot} & \textbf{2-way 8-shot} & \textbf{5-way 8-shot} \\
\hline
\multirow{2}{3cm}{Heavy cold, and sore throat coughing} & TC-ResNet & \textbf{0.79} $\pm$ 0.04 & 0.74 $\pm$ 0.03 & 0.79 $\pm$ 0.04 \\
& ResNet-Attention & 0.76 $\pm$ 0.04 & \textbf{0.85} $\pm$ 0.02 & \textbf{0.81} $\pm$ 0.03\\ 
\hline
\multirow{2}{3cm}{Night wet cough} & TC-ResNet & \textbf{0.77} $\pm$ 0.04 & 0.74 $\pm$ 0.03 & 0.77 $\pm$ 0.03\\ 
& ResNet-Attention & 0.75 $\pm$ 0.02 & \textbf{0.83} $\pm$ 0.02 & \textbf{0.85} $\pm$ 0.02 \\  
\hline
\end{tabular}%
}
\end{center}
\end{table*}

\subsubsection{Comparative study among F2LCough and other approaches  (Q2)}

Table \ref{tab:8-types} shows the performance comparison between our proposed F2LCough and all competitive schemes following the cross-validation approach, including MoCo, Relational network, TC-ResNet (few-shot learning only), ResNet-Attention, TC-ResNet-F (few-shot learning combined with federated learning), and F2LCough under the 2-way 2-shot setting.\footnote{In our experiments, other FSL settings show similar trends.} We summary observations as follows:
\begin{itemize}
\item 
Overall, F2LCough demonstrates effectiveness across most types of cough with superior performance than those of other methods, with the exception of \textit{Dry afternoon cough} and \textit{Gaggy wet cough}.  \item The two types of cough, including \textit{Dry afternoon cough} and \textit{Gaggy wet cough}, pose challenges due to their susceptibility to confusion with other types. 
For the \textit{Dry afternoon cough} case, models trained on multiple devices, such as TC-ResNet-F and F2LCough, exhibit slow or unstable convergence, leading to inferior performance compared to ResNet-Attention and TC-ResNet. 
\item The MoCo approach is the least effective for classifying coughs with all F1-Scores under 0.5. The reason for immensely low performance is that MoCo utilizes a contrastive learning method that requires a significant amount of training data to acquire a good result. 
\item When using a Relational Network the F1-Scores fluctuate between the range of 0.43 and 0.57. One can see that the family of methods following the prototypical approach such as TC-ResNet outweighs the relational network both in the prediction of \textit{Heavy cold, and sore throat coughing} and \textit{Night wet cough}. Via our empirical experiments, we believe that the coughing data tend to form geometric clusters, thus the distance metric employed in the prototypical approach is more appropriate than the relational metric used in the Relational network for the cough classification task. 
\item The F1-scores obtained from methods utilizing the F2L setting are significantly higher than those without federated learning. According to the experimental results, leveraging federated learning in cough classification is one of the most suitable manners, which is capable of exploiting as much user data as possible to preserve user privacy and acquire high performance. Consequently, the federated learning scheme is well-suited for applications, which require preserving user privacy and highly accurate precision like health care applications. 
\end{itemize}
It is worth noting that since federated learning is employed for F2LCough, we only synchronize weights of models of local devices without sharing data in each communication round. As a result, our framework can protect the sensitive data of each user.

\begin{table*}[t]
\caption{The performance comparison of competitive approaches in the 2-way 2-shot few-shot setting}
\label{tab:8-types}
\begin{center}
\addtolength{\tabcolsep}{-5pt}
\resizebox{.9\textwidth}{!}{%
\begin{tabular}{ |c|c|c|c|c|c|c| }
\hline
\textbf{Label} & \textbf{MOCO} & \textbf{\makecell{Relational \\ network} } & \textbf{TC-ResNet} & \textbf{ResNet-Attention} & \textbf{TC-ResNet-F} & \textbf{F2LCough} \\
\hline
{\makecell{Heavy cold, and sore \\ throat coughing}} & 0 
 & 0.52  & 0.79 $\pm$ 0.04 & 0.73 $\pm$ 0.04 & 0.84 $\pm$ 0.04 & \textbf{0.87} $\pm$ 0.02\\
\hline
{Night wet cough} & 0.22 & 0.57 & 0.77 $\pm$ 0.04 & 0.75 $\pm$ 0.02 & 0.82 $\pm$ 0.04 & \textbf{0.85} $\pm$ 0.02\\
\hline
{Dry afternoon cough} & 0.16
 & 0.5 & 0.61 $\pm$ 0.05 & 0.62 $\pm$ 0.03 & \textbf{0.63} $\pm$ 0.03 & 0.57 $\pm$ 0.04
\\
\hline
{Gaggy wet cough} & 0.26 & 0.51 & 0.72 $\pm$ 0.03 & \textbf{0.75} $\pm$ 0.02 & 0.66 $\pm$ 0.04 & 0.67 $\pm$ 0.03\\
\hline
{Spring allergy coughing} & 0.47 & 0.53 & 0.71 $\pm$ 0.02 & 0.66 $\pm$ 0.03 & 0.72 $\pm$ 0.05 & \textbf{0.95} $\pm$ 0.01\\
\hline
{Coughing up crap again} & 0.16	& 0.52 & 0.67 $\pm$ 0.03 & 0.48 $\pm$ 0.04 & 0.68 $\pm$ 0.06 & \textbf{0.95} $\pm$ 0.01
\\
\hline
{Chesty and wet cough} & 0.27 &	0.43 &	0.65 $\pm$ 0.04 &	0.53 $\pm$ 0.1 &	0.78 $\pm$ 0.04 &	\textbf{0.85} $\pm$ 0.02 \\
\hline
{Barking cough} & 0.18 & 0.49 &	0.59 $\pm$ 0.05 &	0.62 $\pm$ 0.05 &	0.78 $\pm$ 0.04 &	\textbf{0.87} $\pm$ 0.02\\
\hline
\end{tabular}%
}
\end{center}
\end{table*}

\subsubsection{Evaluation on real-world scenarios using low-capacity devices (Q3)}

\begin{figure*}[t!]
    \begin{center}
    \includegraphics[width=.80\textwidth]{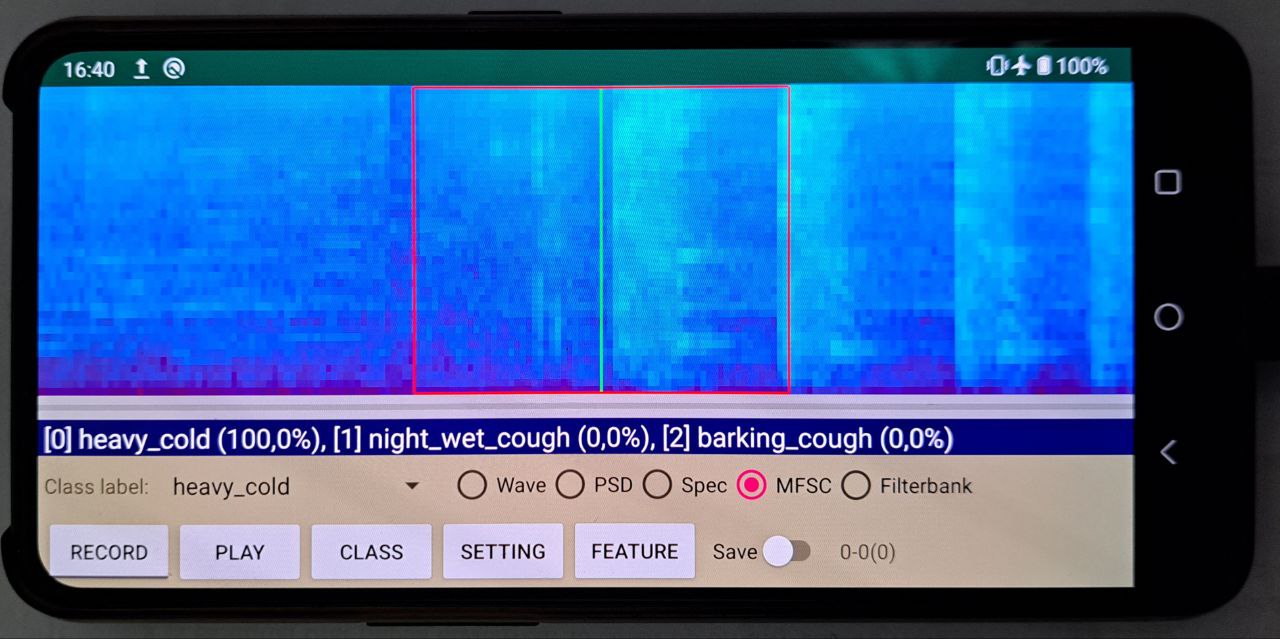}
    \caption{F2LCough application interface}
    \label{fig:interface}
    \end{center}
\end{figure*}

 In a practical scenario, we aim to use the trained model to predict a newly outbreak decease of patients. Thus, the medical centers should be able to deploy the learned model into lightweight devices for field operations. Therefore, to demonstrate the capability of implementing F2LCough in a real-world situation, we experiment on a setup where users record their cough data and save the trained model using an application installed on Android phones while the medical center utilize laptops as mobile devices for training. The interface of the application is shown in Fig. \ref{fig:interface}.

We measure the prediction time of the model integrated on the laptop, shown in Table \ref{tab:running time}. Note that the following laptop specification is used in this experiment:
\begin{itemize}
    \item Laptop (using CPU to train the model): Ram 8GB, Processor 11th Gen Intel(R) Core(TM) i5-1135G7 @ 2.40GHz
\end{itemize}

\begin{table*}[h]
\caption{Time to average and update weights}
\label{tab:running time}
\begin{center}
\scriptsize
\begin{tabular}{ |c|c| }
\hline
\textbf{Type} & \textbf{Running time (ms)}  \\
\hline
Average weights & 184.68  \\
\hline
Update weights & 375.33\\
\hline
\end{tabular}
\end{center}
\end{table*}

Furthermore, we assess the running time on users' Android devices by implementing the F2LCough model using TensorFlow Lite. This implementation is designed to classify various cough sounds, as well as non-cough sounds. The application exhibits a highly efficient inference time, with an average of approximately {\em 2.5 milliseconds}.

To validate the application, we conducted a study with 55 volunteers who tested the application and provided their opinions. The volunteer group consisted of individuals aged 21 to 30, including 5 individuals who work in medical fields. Feedback was collected through a Google form, which included a user satisfaction score ranging from 1 to 5, with 5 representing exceptional satisfaction and 1 indicating a poor experience.

\begin{figure*}[t!]
    \begin{center}
    \includegraphics[width=.80\textwidth]{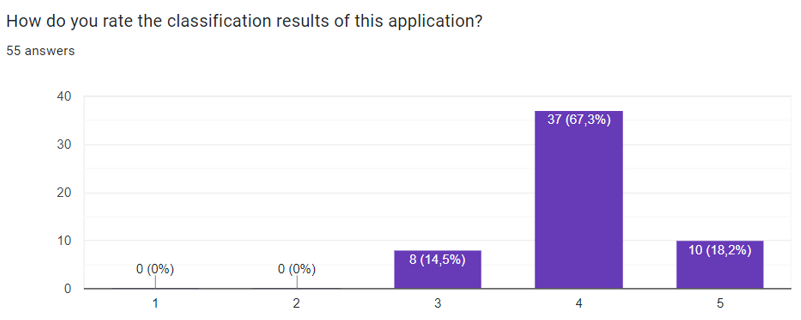}
    \caption{F2LCough application feedback}
    \label{fig:feedback}
    \end{center}
\end{figure*}

The results of the study are depicted in Fig. \ref{fig:feedback}, which displays the user satisfaction statistics. As illustrated, the majority of users rated the application at a score of 4, accounting for 67.3\% of the total responses. Additionally, 18.2\% of users rated the results at a score of 5, while 14.5\% rated the results at a score of 3.
\section{Conclusion}
\label{sec:conclusion}
In this study, we proposed a framework called F2LCough, which is a combination of FSL with federated learning to classify cough sounds. Specifically, we built F2LCough upon ResNet-18 architecture with an attention mechanism and designed a federated learning procedure to learn F2LCough. F2LCough was empirically proved to be an effective method in data scarcity situations and also was able to classify novel data not seen in the training process while still preserving data privacy. Specifically, we obtained 0.87 $\pm$ 0.02 and 0.85 $\pm$ 0.02 average F1-Score for \textit{Heavy cold, and sore throat coughing} and \textit{Night wet cough}, which act as the novel cough types not seen in the training process. Additionally, we also evaluated F2LCough on edge devices such as mobile phones and laptops and found that it exhibited a fast inference time, demonstrating the feasibility of our proposed method.
\section{Acknowledgement}
This work has been supported by the research project coded DT. 17/23, funded by the Ministry of Information and Communication, 2023.

\section*{Declarations}
This section presents declarations related to this study.
\begin{itemize}
\item Funding: This work has been supported by the research project coded DT. 17/23, funded by the Ministry of Information and Communication, 2023. 

\item Conflict of interest/Competing interests: Not applicable

\item Ethics approval: Consent was obtained from all participants prior to their involvement in the study, and they were informed of their right to withdraw at any time without consequence.

\item Consent to participate:
All authors agreed to participate in the construction and development of this research topic.

\item Consent for publication:
All authors agreed to make this study public

\item Availability of data and materials:
The paper uses public data obtained from \cite{ward2021flunet}. The use of data in this study follows the guidelines given by the dataset's authors.

\item Code availability:
Reference sources for building experiments have been mentioned in this paper. We have also included our code repository for research and reference. The source code of our proposed framework is available online at \href{https://github.com/ngandh/F2LCough}.

\item Authors’ contributions:
All authors contributed to the study's conception and design. Cuong Pham and Cong Tran had the idea for the article. Material preparation, data collection, and analysis were performed by Ngan Dao. Experiments were conducted by Ngan Dao and Dat Tran. The first draft of the manuscript was written by Ngan Dao and all authors commented on previous versions of the manuscript. All authors read and approved the final manuscript.

\end{itemize}

\bibliography{sn-article}


\end{document}